\documentclass[times,twocolumn,final,authoryear]{elsarticle}

\usepackage{prletters}
\usepackage{framed,multirow}

\usepackage{amssymb}
\usepackage{latexsym}
\usepackage{subfig}

\usepackage{url}
\usepackage{xcolor}
\definecolor{newcolor}{rgb}{.8,.349,.1}

\journal{Pattern Recognition Letters}

\begin{document}

\thispagestyle{empty}

\clearpage
\thispagestyle{empty}

\ifpreprint
  \vspace*{-1pc}
\else
\fi

\clearpage

\ifpreprint
  \setcounter{page}{1}
\else
  \setcounter{page}{1}
\fi

\begin{frontmatter}

\title{Marrying Tracking with ELM: A Metric Constraint Guided Multiple Feature Fusion Method}

\author[1]{Jing \snm{Zhang}}
\ead{zhangjing_0412@163.com}

\author[2]{Yonggong \snm{Ren}\corref{cor1}}
\cortext[cor1]{Corresponding author:
  Tel.: +86-0411-85992418;
  fax: +86-0411-85992418;}
\ead{ryg@lnnu.edu.cn}

\address[1]{School of computer and information technology, Liaoning Normal University, Liushu south street NO.1 Ganjingzi district, Dalian and 116018, China}

\received{1 May 2013}
\finalform{10 May 2013}
\accepted{13 May 2013}
\availableonline{15 May 2013}
\communicated{S. Sarkar}

\begin{abstract}
Object Tracking is one important problem in computer vision and surveillance system. The existing models mainly exploit the single-view feature (i.e. color, texture, shape) to solve the problem, failing to describe the objects comprehensively. In this paper, we solve the problem from multi-view perspective by leveraging multi-view complementary and latent information, so as to be robust to the partial occlusion and background clutter especially when the objects are similar to the target, meanwhile addressing tracking drift. However, one big problem is that multi-view fusion strategy can inevitably result tracking into non-efficiency.  To this end,
we propose to marry ELM (Extreme learning machine) to multi-view fusion to train the global hidden output weight, to effectively exploit the local information from each view. Following this principle, we propose a novel method to obtain the optimal sample as the target object, which avoids tracking drift resulting from noisy samples. Our method is evaluated over 12 challenge image sequences challenged with different attributes including illumination, occlusion, deformation, etc., which demonstrates better performance than several state-of-the-art methods in terms of effectiveness and robustness.

\end{abstract}


\end{frontmatter}


\section{Introduction}
\label{sec1}
With the rapid development of mobile Internet, multi-media and high performance sensors, object tracking as an important method of image processing has been applied in computer vision and surveillance system. However, object tracking suffers from challenging problems due to image sequences of surveillance usually containing little clarity and fuzzy, which are caused by climate situation, illumination, occlusion, etc. Therefore, how to enhance the tracking performance becomes a hot issue with focusing. Object tracking includes three steps: target description, target detection, target optimization. First, in order to reduce the influence of redundancy and noisy information, the feature extracting method is used to describe the target object including histogram equalization [1], gray level transformation [2], retinex [3] and so on. Histogram equalization expresses color features of images and the gray value distribution effectively. However, the histogram feature has less sensibility to rotate, motion, zoom. Bala, etc. [4] proposed structure element method to express texture feature achieving better performance. However, the information between extracted interest points and image is mismatching in this method, and the structure element contains a strong dependence for images, therefore, the description of this research is limited. In order to obtain the global describe of images, Revaud etc. [5] take advantage of geodesic distance of real interpolation to extract shape feature of image. However, the dimensional of the extracted feature is rather high, and products redundant information. For this problem, multi-scale feature extraction based on compressed sensing had been developed [6, 7]. The method that utilizes relatively small randomly generated linear mapping achieves more accurate the original information of images. Zhang, etc. [8] is inspired by compressed sensing, and proposed CT method. CT method not only fully consideration tracking drift problem, but also well keep the original structure of image. Therefore, this method obtains better performance in target tracking. Meanwhile, Zhang, etc. improved CT model to present FCT model that employs non-adaptive random projections that preserve the structure of the image feature space of objects and compress sample images of the foreground target and the background using the same sparse measurement matrix. This method reduces the computational complex and obtains the real-time tracking performance [9]. However, above methods utilize single-view feature, which result in difficult to fully describe samples. As shown in Fig.1, the object 'Car1' is similar to 'Car2' in shape feature, and the object 'Car1' is similar to 'Car2' in texture feature, and the object 'Car2' is similar to 'Car3' in color feature. In this situation, if any obstruction appears in $t$ frame and the single-view feature incomplete expresses the target object, will produce the important tracking drift problem in $t+1$ frame.
\begin{figure}[!t]
\centering
\includegraphics[width = 8cm]{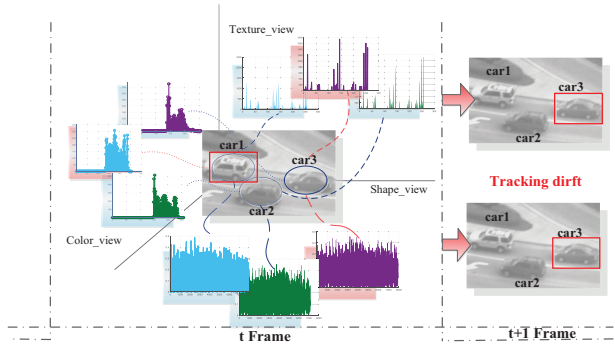}
\caption{Single-view feature leading to tracking drift} \label{fig_sim}
\end{figure}
\par Second, target tracking are continuous and dynamic, the target object at the current frame generates the influence to follow-up tracking directly. Therefore, achieving the suitable detecting method contributes to avoid tracking drift rather effectively. The detecting methods based on discriminative models define the target tracking as binary-classification problem, and obtain the classifier that is used at the next frame. Avidan etc. [10] exploited SVM joining on the optical flow method to in tracking. Comparing SVM, Boosting is a simple and fast method to construct the classifiers relatively. Parag, etc. [11] established the tracking model based on Boosting framework. Zhang, etc. regarded Naive Bayes as the classifier in tracking detection [8]. Moreover, Zhang, etc. [12] proposed STC model that formulates the spatio-temporal relationships between the interested object and its locally dense contexts in the Bayesian framework to solve target ambiguity effectively. Grabner, etc. [13] proposed the online-updating Adaboost algorithm that is used in tracking. Unsupervised classification methods are widely applied in the image processing field due to without the help of any the label information. Wang, etc. [14-27] proposed representative studies including. Kalal, etc. [30] introduced the semi-supervised Adaboost model in the tracking framework, which improves the tracking performance dramatically. Above the tracking detection methods based on discriminative models enhance the tracking performance by optimizing the classification accuracy. Recently, the deep learning model have been proposed and applied in face recognition, segment and so on. Li, etc. [31] took advantage of the CNN model to solve target tracking. Zhou, etc. [32] exploited deep network composing of online boosting to enhance the tracking performance. Above methods based on deep learning exploit only one image at the firstly frame to construct training dataset would lead to the size of training dataset that is small and appear the problem of under-fitting. To address above problems, Hunag, etc. [33, 34] proposed ELM model that is single hidden level neural network. This model convert solve-iterate into linear equation solving by random setting hidden parameters. Comparing deep learning model, this method contains simpler network structure, without a large number of the training dataset, thus enhances the speed of solving and avoid trapping into local optimal solution. Zhang, etc. [34] utilized the incremental learning model of ELM to enhance the tracking robustness and effectiveness. However, the previous models achieve classification for feature extracted samples, and ignore that class labels can provided information for target descriptions.
\par For above problems including inadequate expression, ignoring discriminant information and inefficiency, we proposed the novel tracking method that implements multi-view features fusion by means of supervised class information in the target description. Moreover, we explore the metric constraint between different views to enhance classifying quality in target detection processing. The model framework is displayed in Fig. 2.
\begin{figure}[!t]
\centering
\includegraphics[width = 8.5cm]{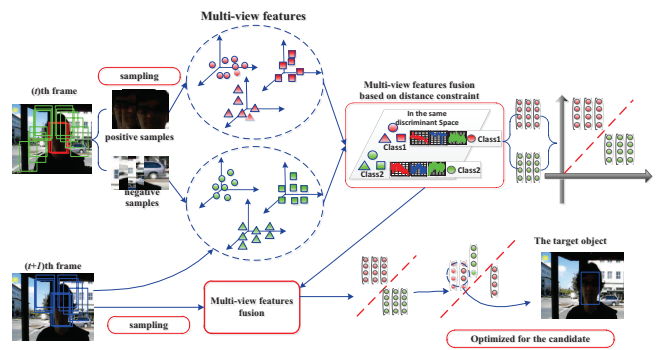}
\caption{ The architecture of the proposed method} \label{fig_sim}
\end{figure}
\par The major contributions are as follows: (1)We present a novel tracking model that achieves the satisfying performance in tracking under special conditions including deformation, occlusion, etc. Meanwhile, it is an incremental tracking approach, so it obtains the real-time tracking; (2)The proposed model fully considers the multi-view attitude of samples, and exploits the multi-view features fusion model to avoid that expression of single-view feature are inaccuracy and incomplete. Meanwhile, introduces the metric constraint to improve the effectiveness of classification in detection and the performance of tracking; (3) In order to enhance the robustness and avoid tracking drift, we presented the novel optimization method to obtain the optimal target object from the candidate set at each frame.
\par This paper is organized as follows. The preliminaries of ELM is introduced in Section 2. The proposed model are detailed in Section 3, and the tracking algorithm and discussion in Section 4. The experimental results are presented in Section 5. The conclusion and our future work in Section 6.
\section{Preliminaries: ELM}
The proposed novel method is based on ELM. In order to facilitate the understanding of our method, this section briefly reviews the related concepts and theories of ELM and developed OSELM.
\par Extreme learning machine is improved by single hidden layer neural network (SLFNs): assume given $N$ samples $(X,T)$, where $X=[x_{1},x_{2},...,x_{N}]^{T} \in\mathbb{R}^{d\times N}$, $T=[t_{1},t_{2},...,t_{N}]^{T} \in\mathbb{R}^{\tilde{N}\times N}$, and $t_{i}=[t_{i1},t_{i2},...,t_{im}]^{T} \in\mathbb{R}^{m}$. The method is used to solve multi-classification problems, and thereby the number of network output nodes is $m(m\geq 2)$. There are $\widetilde{N}$ hidden layer nodes in networks, and activation function $h(\cdot)$ can be Sigmoid or RBF: $\sum_{i=1}^{\widetilde{N}} \beta_{i}h(a_{i}x_{j}+b_{j})=o_{j}$
where $j=1,\cdots,\tilde{N}$, $a_{j}=[a_{j1},a_{j2},\cdots,a_{jd}]^{T}$ is the input weight vector, and $\beta_{j}=[\beta_{j1},\beta_{j2},...,\beta_{jm}]^{T}$ is the output weight vector. Moreover, $a_{j}$, $b_{j}$ can be generated randomly, which is known by. Written in matrix form: $H\beta =T$, where $H_{i} = [h_{1}(a_{1}x_{1}+b_{1}),\cdots, h_{N}(a_{\widetilde{N}}x_{1}+b_{\widetilde{N}})]$. Moreover, the solution form of $H\beta=T$ can be written as: $\hat{\beta}=H^{\dag}T$, where $H^{\dag}$ is the generalized inverse matrix of $H$. ELM minimize both the training errors and the output weights. The expression can be formulated based on optimization of ELM:
\begin{equation}
\begin{array}{c}
  \rm Minimize: \it L_{ELM}=\frac{1}{2}\|\beta\|^{2}_{2}+C\frac{1}{2}\sum_{i=1}^{N} \|\xi_{i}\|^{2}_{2}\\
  \\
  \rm Subject \; to: \it t_{i} \beta\cdot h(x_{i})\geq 1-\xi_{i},i=1,...,N\\
  \\
  \xi_{i}\geq 0,i=1,...,N
\end{array}
\end{equation}
where $\xi_{i}=\left(
                 \begin{array}{ccc}
                   \xi_{i,1} & \cdots & \xi_{i,m} \\
                 \end{array}
               \right)
$ is the vector of the training errors. We can solve the above equation based on KKT theory by Lagrange multiplier, and can obtain the analytical expression of the output weight: $\hat{\beta} = H^{T}(\frac{I}{C}+HH^{T})^{-1}T$. The output function of ELM is: $f(x)=h(x)\hat{\beta}=h(x)H^{T}(\frac{I}{C}+HH^{T})^{-1}T$.
\section{Multi-view feature fusion and discriminative model optimization}
In order to utilize the complementary information and latent knowledge of multi-view feature of sampling sample to improve the performance of tracking, we proposed the novel tracking method based on multi-view features fusion and discriminative model optimizing, including three steps. In describing and predicting of our method, first, achieving the training error under the single-view feature, and the multi-view features are average fused to solve the hidden output weight and the sum of training error. Second, utilize the rate between of the single-view training error and multi-view training error to update the way of multi-view fusion. Finally, minimize the multi-view training error by iterating to obtain the optimal hidden output weight and the fusion way of multi-view feature. We introduce all single-view information when solve the sum of training error, considering the local information of each single-view feature apart from the global information of multi-view feature. Moreover, the metric constraint is utilized to meet that minimize the distinction of the common class and difference views, maximizing the distinction of the common view and difference classes.
\subsection{Features fusion based on minimizing loss of discriminative}
In target detection processing, ELM can be used to classify between the target object and background samples. However, when the sample contains various features, the ELM model is difficult applied to solve the multi-features of a same sample. In this paper, the features fusion method is proposed for this problem. It takes full advantage of multi-view features in tracking to improve the expression of samples and the detection performance of the target object by means of the discrimination information.
\par Given $N$ the different samples, which contain $V$ features of each sample collected in two ways. The feature $v$ corresponds to the samples are: $(x_i^{(v)},t_i^{(v)})$, where $ {X^{(v)}} = {\left[ {x_1^{(v)}, \ldots ,x_N^{(v)}} \right]^T} \in\mathbb{R}{^{{D} \times N}}$, $t_i^{(v)} = {\left[ {t_{i1}^{(v)}, \ldots ,t_{im}^{(v)}} \right]^T} \in\mathbb{R}{^m}$. Meanwhile, the same sample corresponds to the same class, then there is: $t_i^{(1)} = t_i^{(2)} = t_i^{(V)}$. The output weight $\beta$ is solved by using the samples that contain combined multi-view features. The optimization equation is as follows:
\begin{equation}
\begin{array}{l}
L = \frac{1}{2}\left\| \beta  \right\|_2^2 + \sum\limits_{v = 1}^V {{k^{\left( v \right)}}} \left( {\sum\limits_{j = 1}^N {\left\| {\varepsilon _j^{(v)}} \right\|_2^2} } \right)\\
s.t.\begin{array}{*{20}{c}}
{}&{\left( {k \cdot {H^{(v)}}} \right) \cdot \beta }
\end{array} = t_i^T - {\left( {\varepsilon _i^{(v)}} \right)^T}\\
\begin{array}{*{20}{c}}
{}&{}&{\sum\limits_{v = 1}^V {{k^{\left( v \right)}}}  = 1\begin{array}{*{20}{c}}
{}&{k > 0}
\end{array}}
\end{array}
\end{array}
\end{equation}
where $k_{j}$ is the combined parameter that corresponds to the single-features. $\xi _i^{(v)} = \left( {\xi _{i1}^{(v)}, \ldots ,\xi _{im}^{(v)}} \right)$ is training error vector that corresponds to feature $v$. $\beta$ is the output weight vector for different feature space in the equation (6). According to the equation (6), the target is to obtain the minimum value of the training error that combine the different feature space. However, according to the equation, the solution of $[{k_1},{k_2}]$ may be (0,1)/(1,0). In this situation, it will degenerates to the single feature model, and other features are failed. Therefore, we introduction the high power factor $r$, and define $r \ge 2$. The equation is improved as follows:
\begin{equation}
\begin{array}{l}
L = \frac{1}{2}\left\| \beta  \right\|_2^2 + \sum\limits_{v = 1}^V {{k^{\left( v \right)}}} \left( {\sum\limits_{j = 1}^N {\left\| {\varepsilon _j^{(v)}} \right\|_2^2} } \right)\\
s.t.\begin{array}{*{20}{c}}
{}&{\left( {k \cdot {H^{(v)}}} \right) \cdot \beta }
\end{array} = t_i^T - {\left( {\varepsilon _i^{(v)}} \right)^T}\\
\begin{array}{*{20}{c}}
{}&{}&{\sum\limits_{v = 1}^V {{k^{\left( v \right)}}}  = 1\begin{array}{*{20}{c}}
{}&{k > 0}
\end{array}}
\end{array}
\end{array}
\end{equation}
According to the equation (7), we obtain the global optimization hidden output weight under the fused multi-view features, however, the local information of the single view is ignored. The equation (7) is updated as follows:
\begin{equation}
\begin{array}{l}
L = \frac{1}{2}\left\| \beta  \right\|_2^2 + \sum\limits_{v = 1}^V {{k^{\left( v \right)}}} \left( {\sum\limits_{j = 1}^N {\left\| {\varepsilon _j^{(v)}} \right\|_2^2} } \right) + \sum\limits_{v = 1}^V {\left\| {{\sigma ^{(v)}}} \right\|_2^2} \\
s.t.\begin{array}{*{20}{c}}
{}&{\left( {k \cdot {H^{(v)}}} \right) \cdot \beta }
\end{array} = t_i^T - {\left( {\varepsilon _i^{(v)}} \right)^T}\\
\begin{array}{*{20}{c}}
{}&{}&{{\sigma ^{\left( v \right)}} = p \cdot {\beta ^{\left( v \right)}} - \beta }
\end{array}\\
\begin{array}{*{20}{c}}
{}&{}&\begin{array}{l}
\sum\limits_{v = 1}^V {{k^{\left( v \right)}}}  = 1\begin{array}{*{20}{c}}
{}&{k > 0}
\end{array}\\
\sum\limits_{v = 1}^V {{p^{\left( v \right)}}}  = 1\begin{array}{*{20}{c}}
{}&{p > 0}
\end{array}
\end{array}
\end{array}
\end{array}
\end{equation}
where $p$ is the coefficient of combination of single-view information.
\subsection{Metric constraint}
In order to strengthen the discrimination and fully utilize the information of multi-view features, we introduce the metric constraint in Eq.(7).

Furthermore, under the common solution space, the assumption with the distance between samples of different views and same classification that smaller is better, and vice-versa. Therefore, the Eq.(8) is improved as follows:
\begin{equation}
\begin{array}{l}
\mathop {\min }\limits_{\beta ,\varepsilon ,\sigma } \begin{array}{*{20}{c}}
{}
\end{array}\frac{1}{2}\left\| \beta  \right\|_2^2 + \sum\limits_{v = 1}^V {{k^{\left( v \right)}}} \left( {\sum\limits_{j = 1}^N {\left\| {\varepsilon _j^{(v)}} \right\|_2^2} } \right)\\
s.t.\begin{array}{*{20}{c}}
{}&{\left( {k \cdot {S^{(v)}}} \right) \cdot \beta }
\end{array} = t_i^T - {\left( {\varepsilon _i^{(v)}} \right)^T}\\
\begin{array}{*{20}{c}}
{}&{}&{\sum\limits_{v = 1}^V {{k^{\left( v \right)}}}  = 1\begin{array}{*{20}{c}}
{}&{k > 0}
\end{array}}
\end{array}
\end{array}
\end{equation}
Under the view $f$, we define $S=D \dot S$, and $D \in {^{N \times N}}$ is the diagonal matrix that is utilized to expand the distance between samples of different classes. Moreover, the value of element $dii$ in $D$ is ${d_{ii}} = \frac{1}{{{{\left( {{h_i} - \bar h} \right)}^T}\left( {{h_i} - \bar h} \right)}}$, where $\bar h = \frac{1}{N}\sum\limits_{i = 1}^N {{h_i}} $.
\par According to the KKT condition, we obtain the display expression of the hidden output weight $\beta$, and have the following results:
\begin{equation}
\begin{array}{l}
\beta  = \left[ {{{(k_1^{}{S^{(1)}})}^T} +  \cdots  + {{(k_f^{}{S^{(2)}})}^T}} \right]\\
{\left[ {I/C + k_1^{}{S^{(1)}}{{(k_1^{}{S^{(1)}})}^T} +  \cdots  + k_f^{}{S^{(f)}}{{(k_f^{}{S^{(f)}})}^T}} \right]^{ - 1}}T
\end{array}
\end{equation}
\par Meanwhile, we obtain factors of $k^r$ and $p$ as follow:
\begin{equation}
{k_v} = \frac{{1/\sum\limits_{i = 1}^N {{\beta ^T}{s_i}} }}{{\sum\limits_{v = 1}^V {\left( {1/\sum\limits_{i = 1}^N {{\beta ^T}{s_i}} } \right)} }},\begin{array}{*{20}{c}}
{}&{{p_v} = \frac{{1/{\beta ^{(v)}}}}{{\sum\limits_{v = 1}^V {\left( {1/{\beta ^{(v)}}} \right)} }}}
\end{array}
\end{equation}
\section{The tracking algorithm and discussion}
In this paper, first, giving the set from each image sequence ${I_t}$ at the $t$th frame, and obtain positive and negative samples by sampling in the specified range. The sampling method meets two conditions: 1) The positive set closer to the target object at the current frame. 2) The negative set faster to the target object at the current frame. Second, extract multi-view features and achieve features fusion based on the metric constraint. Moreover, obtain the optimal hidden output weight and the candidate target set. Third, we obtain the best sample as the target object at the current frame from the candidate set. Because, noise samples need is avoid, which produces the tracking drift. Therefore, the selected samples is far from the classification boundary as the target object. From the equation (9), the matrix $T$ can be utilized to judge the position relationships between testing samples and the classification boundary. Specific steps are summarized as follows: 1) Calculate the maximum value of each row of $T$ by operation $tmaxV_{i} = max(t_{i})$, and the maximum vector $maxV$ from all of testing samples can be obtained, where $maxV = [tmaxV_{1}, tmaxV_{2},\ldots, tmaxV_{N} ]^{T}\in\mathbb{R}^{N}$. 2) Sort $maxV$ by using the function $smaxV = sort(maxV)$ from the least to the greatest. 3) Obtain the sample sequence $xS$ from testing samples in accordance with elements order of $smaxV$. We define \emph{Former-samples}. There are ranked ahead in the $xS$ sequence, and are close to the classification boundary more. Meanwhile, defined  \emph{Latter-samples} are ranked behind in the $xS$ sequence, and are far from the classification boundary more. Therefore, we should choose \emph{Latter-samples} as candidates that are \emph{Better Target} in target detecting.
\par In this paper, we proposed the novel method to enhance the accuracy in tracking. Meanwhile, in order to fulfill the realtime effectively, it learns the fusion coefficient of multi-view features and the optimal hidden output weight.
\par (1) Difference with related work. Our method based on discriminating model in tracking, comparing to other tracking methods based on discriminative models, such as OAB, SemiB, CT, FCT, SCT, Struct, our method sampling utilizes aggregate and complementary information from difference views, which enhances the performance in samples expression to adapt changing of background. Meanwhile, we introduce the metric constraint to enhance the classification accuracy in target detecting. The proposed method obtains the better performance than OAB and SemiB. Because, OAB and SemiB utilize Adaboost as classifier, which the iterative computation increases consuming of time. And then, OAB exploits only one sample as the positive sample, so it is difficult to sufficient training of discriminating. Moreover, if the target object is the noisy sample, will results from important tracking drift.
\par (2) Robustness to ambiguity in detection. Because tracking is different from traditional binary classification problem, it results from the ambiguity problem easily in the multi-positive samples tracking methods. To address the above problem and enhancing the robustness of the model, we uses the exponential expression of 2-norm of $maxV$ to obtain the optimal sample as the target object. We seek the sample that is far away the classification boundary as the target object, and 2-norm of $maxV$ correlated with the distance from samples to the boundary positively. The proposed method exploits the simple operation to optimize the target object, therefore in ensuring real-time tracking with the premise of the robustness.
\par (3) Robustness to occlusion. In order to distinguish the target object from occlusion, the proposed method fuses multi-view features to fully expression samples and utilizing labels information enhance the performance. Moreover, utilizing the metric constraint of difference views in same classification enhance the discriminant accuracy in target detecting. However, traditional tracking detection methods are difficult to recognize the target object when it is occluded by background image due to lack the distinction between samples description.

\section{Experimental Results}
The proposed algorithm is run on an i7 Quad-Core machine with 3.4GHz CPU, 16GB RAM and a MATLAB implementation. The average efficient of our method is 15 frame per second(FPS), so this method can achieve the tracking effect in real-time. In order to analyze and compare the strength and weakness of the proposed method comprehensively, we exploit 12 image sequences with different attributes to verify the proposed method. Table 1 summarizes all sequences in detail, where attributes record challenges in tracking including low resolution (LR), in$\backslash$out-plane rotation (I$\backslash$OPR), scale variation (SV), occlusion (OCC), deformation (DEF), background clutters (BC), illumination variation (IV), motion blur (MB), fast motion (FM), and outof-view (OV). In addition, we compare the performance with popular tracking methods including OAB, SemiB based on Adaboost classification model, CT, FCT, STC, and comparisons are summarized in Tab. 2, Fig. 4, 5.
\begin{table}
\caption{Description of image sequence in experiments. }
\label{tab:1}       
\begin{tabular}{lllll}
\hline\noalign{\smallskip}
Sequence& Frame & Number & Attributes\\
\noalign{\smallskip}\hline\noalign{\smallskip}
Boy & 602 & 480*640 & I$\backslash$OPR, SV, MB, FM\\
Coke & 291 & 480*640 & I$\backslash$OPR, OCC, IV, FM \\
Couple & 140 & 240*320 & SV, DEF, BC, FM \\
Cardark & 393 & 240*320 & BC, IV \\
David3 & 252 & 480*640  & OPR, OCC, DEF, BC \\
Fish & 476 & 240*320 & IV \\
Girl & 500 & 96*128 & I$\backslash$OPR, SV, OCC \\
Junmming & 75 & 226*400 & MB, FM \\
Mountbike & 228 & 360*640  & I$\backslash$OPR, BC \\
Shaking & 365 & 352*624 & I$\backslash$OPR, BC, SV, IV \\
Singer & 351& 352*624 & OPR, SV, OCC, IV \\
Trellis & 569 & 240*320 & I$\backslash$OPR, BC, SV, IV \\
\noalign{\smallskip}\hline
\end{tabular}
\end{table}
\subsection{Experimental Setup and Evaluation Metrics}
According to the description of the sampling process in Sect. 4, the positive range is $[-10, 10], [-10, 10]$, and the negative range is $[-70, 70], [-70, 70]$. The sliding window is 1. In the detecting process, the number of the hidden-layer node is 300 in our model. Moreover, we extract color, texture, shape features from the common sample to compose the multi-view expression, and PCA model is used to obtain the commonly dimension.
\par Using two kinds of measures evaluate performances between our method and state-of-the-art tracking methods. One way is to calculate the intersecting area, and the computing formula of the success plot is as follow: $S = Area(B_T \cap B_G)/Area(B_T\cup B_G)$, where $B_T$ is the tracked bounding box and denotes $B_G$ is the truth ground. The success plot shows the percentage of frames with $S>t_0$ throughout all threshold $t_0\in [0, 1]$. The other way is to obtain the squared error (SSE) value that is obtained by calculating the difference of the area between the calculated target and the correct target, and the formula is as follow: $SSE = \sqrt {(rx_i-gx_i)^2+(ry_i-gy_i)^2}, i = 1,2, \cdots, N$, where $(rx_i, ry_i)$ is the location of the correct target, and the calculated location is $(gx_i, gy_i)$, and $N$ is the total frames.
\subsection{Tracking results}
\begin{table}
\caption{Success rate(\%)red fonts indicate the first-best tracking performance, and green fonts indicate the second-best tracking performance. }
\label{tab:1}       
\begin{tabular}{llllllll}
\hline\noalign{\smallskip}
       &Ours & OAB & SemiB & STC & CT  & FCT & Struck \\
\noalign{\smallskip}\hline\noalign{\smallskip}
Boy & {\color{red}80} & {\color{green}73} & 21 & 45 & 48  & 72 & 51 \\
Coke & {\color{red}39} & {\color{green}19} & 0 & 1 & 2  & 1 & {\color{green}19} \\
Couple & {\color{red}60} & 28 & 14 & 0 & 40  & {\color{green}56} & 12 \\
Cardark & {\color{red}78} & 35 & 26 & 18 & 63  & 66 & {\color{green}76} \\
David3 & {\color{red}55} & 42 & {\color{green}43} & 13 & 32  & 10 & 21 \\
Fish & {\color{red}93} & 54 & 1 & 67 & 50  & {\color{green}90} & 72 \\
Girl & {\color{red}59} & 1 & 29 & 1 & 2  & 27 & {\color{green}44} \\
Junmming & {\color{red}61} & 32 & 17 & 29 & 56  & 37 & {\color{green}47} \\
Mountbike & {\color{green}88} & 63 & 18 & 59 & 18  & {\color{red}86} & 33 \\
Shaking & {\color{red}80} & 10 & 0 & 0 & {\color{green}17}  & 1 & 1 \\
Singer & {\color{red}75} & 18 & 15 & 22 & 22  & {\color{red}74} & {\color{green}63} \\
Trellis & {\color{red}20} & {\color{green}1} & 0 & 0 & 0  & 0 & {\color{green}1} \\
\noalign{\smallskip}\hline
\end{tabular}
\end{table}
(1) Overall performance: table 2 summarizes the overall performance of challenging sequences in terms of precision plots, where red fonts indicate the first-best tracking performance in each row of table 2, and green fonts indicate the second-best tracking performance. Meanwhile, OAB, SemiB, STC, CT, FCT, Struck are provided by the author. From results in table 2, we note that our method achieves the first-best performance in most sequences causing by sufficiently representing each sampling image on tracking. Moreover, the proposed method obtains the optimal discriminative model based on multi-view features fusion. Therefore, our model have satisfying tracking performance, in particular, it obtains the robustness on Suv, Coke, Trellis that are occluded by background. However, FCT attain the first-best tracking performance on Dog, Mountbike, Singer sequence, and CT attain the first-best tracking performance on Duke sequence. OAB and SemiB exploit the Adaboost classification model in detecting process. However, SemiB has the tracking effect is not ideal due to use a lot of unlabel simples to build the discrimination model. STC, CT and FCT utilize the Navie Bayes classification model in tracking detection process. CT and FCT build the low dimension expression and retain important features from original samples, therefore, achieve the second-best performance in challenging sequences. The real-time is an important term of tracking. Our method is an incremental tracking model that only learns the current image and need not to learn the cumulate data in bulk. CT and FCT get the low dimension expression by random matrix, thereby enhance the computation speed. Therefore, FCT is the most effective tracking model among all algorithms. However, in SemiB, the tracking performance is not ideal due to a lot of unlabel samples participating in building the discrimination model. Struck achieves the solution by iterating in the learning process, thus impacts the effectiveness.

\section{Conclusion}
In this paper, we proposed an efficient and robust tracking method based on ELM. The proposed method exploits fused multi-view feature to enhance expression of samples and effectiveness of detecting, which achieve the satisfying tracking performance. Meanwhile, we proposed the optimization method to obtain the optimized target object from the candidate set, thereby avoids the tracking drift problem that is caused by noisy samples. We utilize 12 image sequences to evaluate the proposed method, which achieves more performance than several state-of-the-art methods in effective and robustness. In addition, we will explore more efficient tracking model using the multi-feature image sequences.

\section*{References}
\begin{enumerate}
\item Schindelin, Johannes, et al. The ImageJ ecosystem: an open platform for biomedical image analysis. Molecular reproduction and development 82.7-8 (2015): 518-529;
\item Adam, Amit, Ehud Rivlin, and Ilan Shimshoni. Robust fragments-based tracking using the integral histogram. Computer vision and pattern recognition, 2006 IEEE Computer Society Conference on. Vol. 1. IEEE, 2006;
\item He, Xiao, et al. Networked strong tracking filtering with multiple packet dropouts: algorithms and applications. IEEE Transactions on Industrial Electronics 61.3 (2014): 1454-1463;
\item Bala A, Kaur T. Local texton XOR patterns: a new feature descriptor for content-based image retrieval[J]. Engineering Science and Technology, an International Journal, 2016, 19(1): 101-112;
\item Revaud, Jerome, et al. Epicflow: Edge-preserving interpolation of correspondences for optical flow. Proceedings of the IEEE Conference on Computer Vision and Pattern Recognition. 2015;
\item Donoho, David L. Compressed sensing. IEEE Transactions on information theory 52.4 (2006): 1289-1306;
\item Candes, Emmanuel J., and Terence Tao. Near-optimal signal recovery from random projections: Universal encoding strategies?. IEEE transactions on information theory 52.12 (2006): 5406-5425;
\item Zhang, Kaihua, Lei Zhang, and Ming-Hsuan Yang. Real-time compressive tracking. European Conference on Computer Vision. Springer Berlin Heidelberg, 2012;
\item Zhang K, Zhang L, Yang M H. Fast compressive tracking[J]. IEEE transactions on pattern analysis and machine intelligence, 2014, 36(10): 2002-2015;
\item Avidan S. Support vector tracking [J]. Pattern Analysis and Machine Intelligence, IEEE Transactions on,2004,26(8): 1064-1072;
\item Parag T, Porikli F, Elgammal A. Boosting adaptive linear weak classifiers for online learning and tracking[C]//Computer Vision and Pattern Recognition, 2008. CVPR 2008. IEEE Conference on. IEEE, 2008: 1-8.
\item Zhang K, Zhang L, Liu Q, et al. Fast visual tracking via dense spatio-temporal context learning[C]//European Conference on Computer Vision. Springer International Publishing, 2014: 127-141.
\item B. Babenko, M.-H. Yang, and S. Belongie. Robust object tracking with online multiple instance learning[J]. IEEE Transactions on Pattern Analysis and Machine Intelligence, vol. 33, no. 8, pp. 1619-1632, 2011;
\item Y. Wang, X. Lin, L. Wu, W. Zhang, Q. Zhang, X. Huang. Robust subspace clustering for multi-view data by exploiting correlation consensus. IEEE Transactions on Image Processing, 24(11):3939-3949,2015.
\item Y. Wang, X. Lin, L. Wu, W. Zhang. Effective Multi-Query Expansions: Collborative Deep Networks for Robust Landmark Retrieval. IEEE Transactions on Image Processing, 26(3):1393-1404, 2017.
\item Y. Wang, W. Zhang, L.Wu, X. Lin, X. Zhao. Unsupervised metric fusion over multiview data by graph random walk-based cross-view diffusion. IEEE Transactions on Neural
Networks and Learning Systems, 28(1):57-70, 2017.
\item Y. Wang, L. Wu, X. Lin, J. Gao. Multiview spectral clustering via structured low-rank matrix factorization. IEEE Transactions on Neural Networks and Learning Systems, 2018.
\item Y. Wang, X. Lin, L. Wu, W. Zhang. Effective Multi-Query Expansions: Robust Landmark Retrieval. ACM Multimedia 2015:79-88.
\item Y. Wang, X. Lin, L. Wu, W. Zhang, Q. Zhang. LBMCH: Learning Bridging Mapping for Cross-modal Hashing. ACM SIGIR 2015.
\item Y. Wang, W. Zhang, L. Wu, X. Lin, M. Fang and S. Pan. Iterative views agreement: An iterative low-rank based structured optimization method to multi-view spectral clustering. IJCAI 2016: 2153-2159.
\item Y. Wang, L. Wu. Beyond low-rank representations: Orthogonal clustering basis reconstruction with optimized graph structure for multi-view spectral clustering. Neural Networks, 103:1-8, 2018.
\item L. Wu, Y. Wang, X. Li, J. Gao. Deep Attention-based Spatially Recursive Networks for Fine-Grained Visual Recognition. IEEE Transactions on Cybernetics, 2018.
\item L. Wu, Y. Wang, J. Gao, X. Li. Deep Adaptive Feature Embedding with Local Sample Distributions for Person Re-identification. Pattern Recognition, 73:275-288, 2018.
\item L. Wu, Y. Wang, X. Li, J. Gao. What-and-Where to Match: Deep Spatially Multiplicative Integration Networks for Person Re-identification. Pattern Recognition, 76:727-738, 2018.
\item L. Wu, Y. Wang, L. Shao. Cycle-Consistent Deep Generative Hashing for Cross-Modal Retrieval. arXiv:1804.11013, 2018.
\item Y. Wang, X. Huang, L. Wu. Clustering via geometric median shift over Riemannian manifolds. Inf. Sci. 220: 292-305 (2013)
\item Y. Wang, X. Lin, L. Wu, Q. Zhang, W. Zhang. Shifting multi-hypergraphs via collaborative probabilistic voting. Knowl. Inf. Syst. 46(3): 515-536 (2016)
\item Z. Kalal, J. Matas, and K. Mikolajczyk. Pn learning: Bootstrapping binary classi?ers by structural constraints[C]// in Proceedings of IEEE Conference on Computer Vision andPattern Recognition, pp. 49-56, 2010;
\item S. Wang, H. Lu, F. Yang, and M.-H. Yang, Superpixel tracking, in Proceedings of the IEEE International Conference on Computer Vision, pp. 1323-1330, 2011
\item H. Li, Y. Li, and F. Porikli, Robust online visual tracking with a single convolutional neural network, in Proc. 12th Asian Conf. Comput. Vis., 2014, pp. 194-209.
\item X. Zhou, L. Xie, P. Zhang, and Y. Zhang. An ensemble of deep neural networks for object tracking[C]//IEEE Int. Conf. Image Process., Oct. 2014, pp. 843-847.
\item Huang G B, Zhu Q Y, Siew C K. Extreme learning machine: theory and applications[J]. Neurocomputing, 2006, 70(1): 489-501.
\item Huang G, Huang G B, Song S, et al. Trends in extreme learning machines: A review[J]. Neural Networks, 2015, 61: 32-48.
\item Zhang J, Feng L, Yu L. A novel target tracking method based on OSELM[J]. Multidimensional Systems and Signal Processing, 2016: 1-18.
\end{enumerate}

\end{document}